\documentclass{article} 

\usepackage{times}
\usepackage{graphicx} 
\usepackage{subfigure} 
\usepackage{multirow}
\usepackage{natbib}
\usepackage{amsmath}
\usepackage{algorithm}
\usepackage{algorithmic}
\usepackage{iclr2017_workshop,times}
\usepackage{hyperref}
\usepackage{url}

\title{Training Triplet Networks with GAN}

\author{Maciej Zieba\\
Department of Computer Science and Management\\
Wroclaw University of Science and Technology\\
Wroclaw, Poland\\
\texttt{maciej.zieba@pwr.edu.pl} \\
\And
Lei Wang \\
School of Computing and \\
Information Technology \\
University of Wollongong \\
Wollongong, Australia \\
\texttt{leiw@uow.edu.au}
}

\begin{document}

\maketitle

\begin{abstract}
Triplet networks are widely used models that are characterized by good performance in classification and retrieval tasks. In this work we propose to train a triplet network by putting it as the discriminator in Generative Adversarial Nets (GANs). We make use of the good capability of representation learning of the discriminator to increase the predictive quality of the model. We evaluated our approach on Cifar10 and MNIST datasets and observed significant improvement on the classification performance using the simple $k$-nn method.            
\end{abstract}

\section{Introduction}

Generative Adversarial Nets (GAN) \citep{GO:14} are one of the state-of-the-art solutions in the field of generative models in computer vision. Recent studies on GANs also show their great capabilities in feature extraction \citep{RA:15,DO:16} and classification tasks \citep{SA:16}. These are mainly achieved by incorporating feature matching technique in training the generator and multitask training of the discriminator, that plays an additional role as a classifier. 

Inspired by the semi-supervised framework described by \cite{SA:16}, we propose a novel model for triplet network learning. The feature layer that is involved in feature matching technique during generative part of the training is further used as a triplet output in supervised part of training the discriminator. As a consequence, the feature representation of the triplet output is enriched by the consequences of GAN's training. This benefit is especially observed when the access to the labeled triplets is limited.   

In the paper, we make the following contributions: 1) we propose the novel method for training triplet network using GAN framework; 2) we show, how to obtain stronger representation from GAN model if we have access to some portion of labeled triplets; 3) we show in the experiment, that with only $16$ features our model has been able to produce competitive classification performance using the simple $k$-nn method.     

\section{Triplet Networks with GAN }
\label{architecture}

\subsection{Triplet Networks}
\label{triplet}

Triplet networks \citep{HO:15} are one of the most commonly used techniques in deep learning metric \citep{YO:16,ZH:16}. The main idea that stays behind them is to take the set of triplets (training data), where each triplet is composed of  query $\mathbf{x}^{q}$ (assumed to be positive), positive $\mathbf{x}^{+}$ and negative $\mathbf{x}^{-}$ examples, and train the network $T(\mathbf{x})$ to construct the effective feature extractor. The model makes use of the probability ($p_{T(\mathbf{x}^{q},\mathbf{x}^{+},\mathbf{x}^{-})}:=p_T$) that the distance of the query example to the negative example is greater than its distance to the positive one: $p_T = \frac{\exp\{{d_T(\mathbf{x}^{q},\mathbf{x^{-}})\}}}{\exp\{{d_T(\mathbf{x}^{q},\mathbf{x^{-}})\}} + \exp\{{d_T(\mathbf{x}^{q},\mathbf{x^{+}})\}}}$, where $d_T(\mathbf{x}_1,\mathbf{x}_2)$ is defined as Euclidean distance of the outputs of $T(\cdot)$: $d_T(\mathbf{x}_1,\mathbf{x}_2) = || T(\mathbf{x}_1) - T(\mathbf{x}_2 )||_2$.

The loss function for a single triplet $(\mathbf{x}^{q}_n,\mathbf{x}^{+}_n,\mathbf{x}^{-}_n)$ can be defined as $L_T = -\log{(p_{T(\mathbf{x}^{q}_n,\mathbf{x}^{+}_n,\mathbf{x}^{-}_n)})}$. We propose to use slightly different lost function from that was defined by \cite{HO:15} because we want to be consistent with \emph{log-prob} discriminative part of learning for GAN model.

\subsection{Generative Adversarial Nets (GAN)}
\label{gan}

The main idea of GANs is based on game theory and assumes training of two competing network structures, generator $G(\mathbf{z})$ and discriminator $D(\mathbf{x})$. The goal of GANs is to train generator $G$ to sample from the data distribution $p_{data}(\mathbf{x})$ by transforming the vector of noise $\mathbf{z}$. The discriminator $D$ is trained to distinguish the samples generated by $G$ from the samples from $p_{data}(\mathbf{x})$. The training problem formulation is as follows:  $\min_{G} \max_{D} V(D,G) = \mathrm{E}_{\mathbf{x} \sim p_{data}(\mathbf{x})}{[\log{(D(\mathbf{x}}))]} +\mathrm{E}_{z \sim p_{\mathbf{z}}(\mathbf{z})}{[\log{(1 - D(G(\mathbf{z})}))]}$, where $p_{\mathbf{z}}(\mathbf{z})$ is prior over $\mathbf{z}$.  

The model is usually trained with the SGD approach by sampling minibatch of fakes from $p_{\mathbf{z}}(\mathbf{z})$  and minibatch of data samples from $p_{data}(\mathbf{x})$. They are used to maximize $V(D,G)$ with respect to parameters of $D$ by assuming a constant $G$, and then minimize $V(D,G)$  with respect to parameters of $G$ by assuming a constant $D$. The procedure is repeated for each of the epochs. 

\subsection{Semi-supervised training with GAN}
\label{ssgan}

The most recent studies on GAN show the great benefit of using them in semi-supervised (and supervised) classification. The main idea of this approach is to incorporate the discriminator $D$ into an additional classification task. As a consequence, $D$ is trained both to distinguish fake and true samples and to classify the examples to one of the predefined classes in the classification. 

The loss function for the discriminator can be defined as a sum of supervised and unsupervised parts, $L_D = L_s + L_{u}$. The supervised part is defined as $L_{s}= -\mathrm{E}_{\mathbf{x},y \sim p_{data}(\mathbf{x},y)}{[\log{(p(y|\mathbf{x})})]}$, where $y$ denotes class label. The unsupervised part of the criterion is defined as $L_u = - V(D,G)$.

To improve the quality of prediction and to obtain better feature representation, \cite{SA:16} recommend, that generator $G$ is trained using so-called \emph{feature matching} procedure. The objective to train the generator $G$ is $L_G = ||\mathrm{E}_{\mathbf{x} \sim p_{data}(\mathbf{x})}\mathbf{f}(\mathbf{x}) - \mathrm{E}_{z \sim p_{\mathbf{z}}(\mathbf{z})}{\mathbf{f}(G(\mathbf{z}))}||_2^2$, where $\mathbf{f}(\mathbf{x})$ denotes the intermediate activation layer of the discriminator.
 
\subsection{Triplet training with GAN}
\label{sstgan}

In our approach we make use of benefits of using triplet networks for metric learning and the effectiveness of semi-supervised GAN in classification tasks. The main idea behind our approach is to incorporate discriminator in a metric learning task instead of involving it in classification. As a consequence, we aim at obtaining a good feature representation in generative part of training, but also in supervised part of training the discriminator. 

We assume the output of the proposed triplet network is characterized by $M$ features, $T(\mathbf{x}) = [t_1(\mathbf{x}),\dots,t_M(\mathbf{x})]^T$. Inspired by \cite{SA:16}, we define the discriminator $D_T(\mathbf{x}) $ in the following manner: $D_T(\mathbf{x}) = \frac{\sum_{m=1}^{M}\exp(t_m(\mathbf{x})) }{\sum_{m=1}^{M}\exp(t_m(\mathbf{x}))+1}$. It indicates the posteriori probability of being real examples, while the posteriori probability for fake examples is just $1 -D_T(\mathbf{x})$. Certainly, we can train an additional layer on top of this feature layer as in common practice. However, we find that this does not give clear advantage over the $D_T(\mathbf{x})$ defined as above, since we essentially only need a mapping from the features to the probability. So this $D_T(\mathbf{x})$ is employed in our model for the sake of efficiency.

The loss function used for training triplet discriminator is composed of triplet-based and unsupervised components $L_{TD} = L_{Ts} + L_{Tu}$. We define the triplet based component as follows $L_{Ts}= -\mathrm{E}_{\mathbf{x}^{q},\mathbf{x}^{+},\mathbf{x}^{-} \sim p_{data}(\mathbf{x}^{q},\mathbf{x}^{+},\mathbf{x}^{-} )}{[\log{(p_T)}]}$ and the unsupervised part remains unchanged $L_{Tu} = - V(D_T,G)$. 

For supervised loss component ($L_{Ts}$) we sample labeled triplets from data distribution $p_{data}(\mathbf{x}^{q},\mathbf{x}^{+},\mathbf{x}^{-} )$. The unsupervised loss component is trained in classical manner using triplet discriminator $D_T$.

The generative part of the model utilizes the feature matching, where the vector of matched outputs represents the output of triplet network, $T(\mathbf{x}) = \mathbf{f}(\mathbf{x})$. The triplet output is further used in $k$-nn-based search with a Euclidean distance.  
  
\section{Experiments}

The goal of the experiment is to evaluate the quality of the triplet model using benchmark datasets. We make use of the structures proposed by \cite{SA:16} and modify the output layer to triplet manner. The number of output features $M$ is set as low as $16$. We use standard data split for the benchmark datasets, $50000$ training and $10000$ test examples (Cifar10), $60000$ training and $10000$ test cases (MNIST). We take only $100$ labeled examples for MNIST, and $5000$ for Cifar10 from their training splits. The weights of our model are initialized using the GAN model pretrained in unsupervised manner. $500$ epochs are used during using MNIST dataset and $700$ epochs for Cifar10.  

\begin{table}[t]
\caption{Classification accuracy on benchmark datasets. $9$-nn is used for evaluation on $16$ features. }
\label{tab}
\begin{center}
 \begin{tabular}{|l|c|c|}
\emph{Method} & \textbf{MNIST} & \textbf{Cifar10} \\
    \hline
  Only Triplet & 81.27 & 70.76 \\
  Only GAN & 96.48 & 55.39 \\
  Our method & 97.50 & 80.97 \\
  \end{tabular}
\end{center}
\end{table}

The number of possible triplets for Cifar10 was large, so we applied simple hard-mining technique before each of the training epochs. For each of $K$ labeled query examples we determine the $N$ positive examples that are, according to model $T$ on current training stage, the most distant to the considered example. We select also $N$ negatives that are the closest to the considered example. As a consequence, the query example, the most distant positive and the closest negative form one triplet. We continue creating the triplets by taking closer positive and more distant negative to obtain $K$ triplets for one query example. Using selected positives and negatives we form the $N\cdot K$ triplets to balance the number of unsupervised data. For MNIST the total number of possible triplets is $90000$, which is not very large. So we simply randomly select $60000$ for each epoch.

The results of initial experiments are presented in Table \ref{tab}. We compared our approach with the triplet network trained only on labeled examples and the GAN model that is trained only on unlabeled data.  The proposed approach outperforms the two methods and the improvement on Cifar10 is significant. In addition, we trained our model using all $50000$ labeled examples for Cifar10 data ($16$ features, $9$-nn classifier) and we obtained the accuracy of classification equal $88.04$. This result is promising comparing to the classification accuracy of triplet network reported by \cite{HO:15} ($87.10$), taking into account that we use only 16 features, $9$-nn classifier, no data augmentation and no models pretrained on external data are performed. Currently, we aim at better model selection and better balance during training procedure to improve the performance in further. We also investigated the case of more features and obtained $98.68$ classification accuracy on MNIST dataset ($256$ features, $9$-nn classifier).

\section{Conclusion}

In this work we present a novel framework for learning triplet models that makes use of GAN model to obtain better feature representation. The presented model can be easily applied for image retrieval tasks, where only a small portion of labeled data could be accessed. This model shows promising results even when the number of features is low, which is computationally desirable for the methods like $k$-nn search. Very recently, we noticed that the paper \citep{AR:17} on improving GAN with Wasserstein distance could be beneficial to our research. We plan to incorporate this approach into our model to improve its performance in further.   

\subsubsection*{Acknowledgments}
This work was undertaken with financial support of a Thelxinoe grant in the context of the EMA2/S2 THELXINOE: Erasmus Euro-Oceanian Smart City Network project, grant reference number: 545783-EM-1-2013-1-ES-ERA MUNDUS-EMA22.

\bibliographystyle{iclr2017_workshop}

\newpage

\section*{Appendix} 

\begin{table}[htb]
\caption{Mean average precision (mAP) values on Cifar10 dataset.}
\label{tab:2}
\begin{center}
 \begin{tabular}{|l|c|c|c|c|}
  & \textbf{Our approach} & \textbf{Only triplet} & \textbf{Only GAN} & \textbf{SCGAN} \\
  \hline
 mAP &  $0.6525$ & $0.5367$ & $0.2003$ & $0.4266$ 
  \end{tabular}
\end{center}
\end{table}

We introduce the additional results for the proposed model. In Table \ref{tab:2} we present the comparison for Cifar10 with respect to mean average precision (mAP) criterion referring to semi-supervised classification GAN (SCGAN) \citep{SA:16}. For the SCGAN we take penultimate layer of the discriminator as feature representation. It can be observed, that our approach outperforms reference solutions.

\begin{table}[htb]
\caption{Classification and retrieval results obtained on MNIST data for increasing number of features ($m$-number of features, $9$-NN is used as classification model, number of labeled examples is equal to $100$)}
\label{tab:3}
\begin{center}
 \begin{tabular}{|l|c|c|c|c|c|}
  & \textbf{m=16} & \textbf{m=32} & \textbf{m=64} & \textbf{m=128} & \textbf{m=256} \\
  \hline
Accuracy  &  $97.61\%$  & $98.26\%$ & $98.31\%$ & $98.69\%$ & $ 98.65\%$ \\
  \hline
mAP & $ 0.8929 $ & $0.9118$ & $ 0.9056 $ & $ 0.9321 $ & $0.9414$ \\
  \end{tabular}
\end{center}
\end{table}

\begin{table}[htb]
\caption{Classification and retrieval results obtained on MNIST data for increasing number of labeled examples ($N$-number of labeled examples, $9$-NN is used as classification model, number of features is equal to $16$)}
\label{tab:4}
\begin{center}
 \begin{tabular}{|l|c|c|c|c|c|}
  & \textbf{N=100} & \textbf{N=200} & \textbf{N=500} & \textbf{N=1000} \\
  \hline
Accuracy  &  $97.61\%$  & $98.50\%$ & $98.59\%$ & $ 98.86\%$ \\
  \hline
mAP & $  0.8929 $ & $0.9244$ & $ 0.9588 $ & $ 0.9700 $ 
  \end{tabular}
\end{center}
\end{table}   

In Tables \ref{tab:3} and \ref{tab:4} we present the classification and retrieval results obtained on MNIST data for different number of features and labeled examples. As seen, the performance of our approach is relatively stable, and it improves with the increasing number of labeled examples and features.  

\begin{table}[ht!]
\caption{Classification and retrieval results obtained on Cifar10 data for increasing number of features ($m$-number of features, $9$-NN is used as classification model, number of labeled examples is equal to $5000$)}
\label{tab:5}
\begin{center}
 \begin{tabular}{|l|c|c|c|c|c|}
  & \textbf{m=16} & \textbf{m=32} & \textbf{m=64} & \textbf{m=128} & \textbf{m=256} \\
  \hline
Accuracy  &  $80.97\%$  & $80.86\%$ & $79.43 \%$ & $77.43 \%$ & $ 76.65 \%$ \\
  \hline
mAP & $ 0.6525 $ & $0.6609$ & $ 0.6438 $ & $ 0.6196 $ & $0.6028$ \\
  \end{tabular}
\end{center}
\end{table}

\begin{table}[ht!]
\caption{Classification and retrieval results obtained on Cifar10 data for increasing number of labeled examples ($N$-number of labeled examples, $9$-NN is used as classification model, number of features is equal to $16$)}
\label{tab:6}
\begin{center}
 \begin{tabular}{|l|c|c|c|}
  &  \textbf{N=2000} & \textbf{N=5000} & \textbf{N=50000} \\
  \hline
Accuracy  &  $78.30\%$  & $80.97\%$ & $88.04\%$ \\
  \hline
mAP & $  0.5783 $ & $0.6525$ & $ 0.8097 $ 
  \end{tabular}
\end{center}
\end{table}

In Tables \ref{tab:5} and \ref{tab:6} we present the classification and retrieval results obtained on Cifar10 data for different number of features and labeled examples. For increasing number of labeled examples we can see the improvement of quality of the model. However, for increasing number of features we observed decreasing quality of the model. This phenomenon is especially observed, when the vector of considered features is composed of $128$ units or more. We diagnosed the reason that stays behind the hard mining technique applied for the dataset. The larger vectors of features have the tendency to overfit to the hard examples at the expense of easy cases. This should have been compensated by dynamic selection of hard mining cases. However, due to the computational cost, the dynamic procedure for selecting hard triplets is currently performed for each epoch, instead of each minibatch update of the weights. In the future work, we will improve  the dynamic procedure for selecting the hard triplets.
\end{document}